\documentclass[10pt,twocolumn,letterpaper]{article}

\usepackage[pagenumbers]{cvpr} 

\usepackage{graphicx}
\usepackage{amsmath}
\usepackage{amssymb}
\usepackage{enumerate}
\usepackage{booktabs}
\usepackage{color, colortbl}
\usepackage[accsupp]{axessibility}

\usepackage[pagebackref,breaklinks,colorlinks]{hyperref}

\usepackage[capitalize]{cleveref}
\crefname{section}{Sec.}{Secs.}
\Crefname{section}{Section}{Sections}
\Crefname{table}{Table}{Tables}
\crefname{table}{Tab.}{Tabs.}

\def\cvprPaperID{2682} 
\def\confName{CVPR}
\def\confYear{2022}

\begin{document}

\title{Democracy Does Matter: Comprehensive Feature Mining for Co-Salient Object Detection}

\author{Siyue Yu\textsuperscript{1,2}, ~~~~~~~ Jimin Xiao\textsuperscript{1}\thanks{corresponding author}, ~~~~~~~ Bingfeng Zhang\textsuperscript{1,2}, ~~~~~~~Eng Gee Lim\textsuperscript{1} \\
{\textsuperscript{1}XJTLU, ~~~~~~~} {\textsuperscript{2}University of Liverpool}\\
{\tt\small \{siyue.yu,jimin.xiao,bingfeng.zhang,enggee.lim\}@xjtlu.edu.cn}
}

\maketitle
\begin{abstract}
Co-salient object detection, with the target of detecting co-existed salient objects among a group of images, is gaining popularity. Recent works use the attention mechanism or extra information to aggregate common co-salient features, leading to incomplete even incorrect responses for target objects.
In this paper, we aim to mine comprehensive co-salient features with democracy and reduce background interference without introducing any extra information.
To achieve this, we design a democratic prototype generation module to generate democratic response maps, covering sufficient co-salient regions and thereby involving more shared attributes of co-salient objects. Then a comprehensive prototype based on the response maps can be generated as a guide for final prediction. 
To suppress the noisy background information in the prototype, we propose a self-contrastive learning module, where both positive and negative pairs are formed without relying on additional classification information.
Besides, we also design a democratic feature enhancement module to further strengthen the co-salient features by readjusting attention values.
Extensive experiments show that our model obtains better performance than previous state-of-the-art methods, especially on challenging real-world cases (\eg, for CoCA, we obtain a gain of 2.0\% for MAE, 5.4\% for maximum F-measure, 2.3\% for maximum E-measure, and 3.7\% for S-measure) under the same settings. Source code is available at \url{https://github.com/siyueyu/DCFM}\footnotetext[1]{The work was supported by National Natural Science Foundation of China under 61972323.}.
\end{abstract}

\section{Introduction}
\label{sec:intro}
\begin{figure}
\centering
	\includegraphics[scale=0.41]{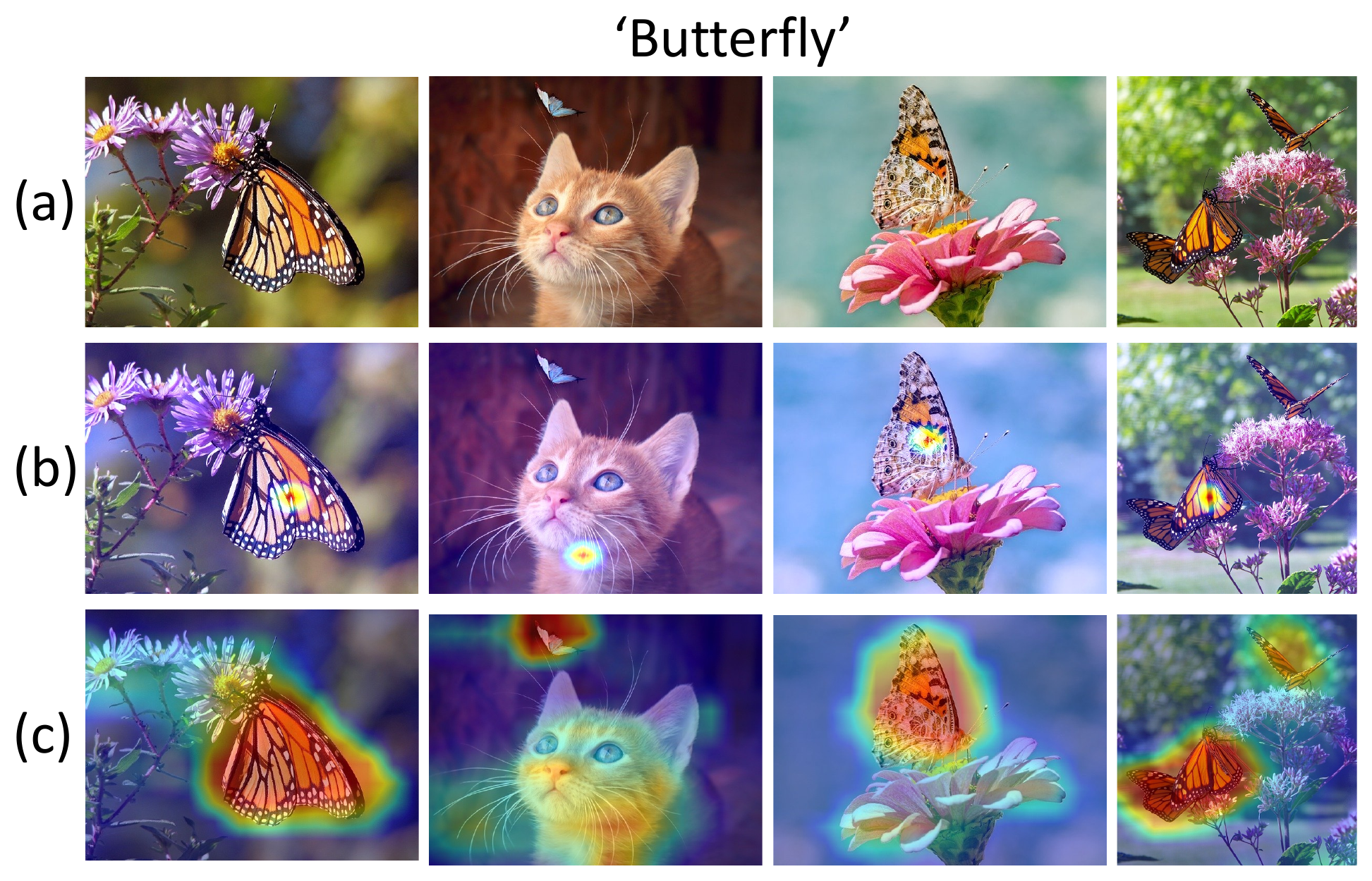}
	\caption{Visualization of response maps. (a) Inputs; (b) Response maps generated by the previous approach~\cite{fan2021group}; (c) Ours. It can be seen that ours can cover more co-salient objects.
	}
	\label{FIG:cormap}
\end{figure}
Co-salient object detection (CoSOD) aims to detect the common salient objects among a group of input images. Unlike salient object detection (SOD), which is to detect the most attractive objects by mimicking human eyes~\cite{chen2020global, li2021uncertainty, yu2021structure, tang2021disentangled, pang2020multi, li2021salient, piao2021mfnet, liu2021samnet}, CoSOD focuses on detecting salient yet co-existed objects among all the input images. In this case, CoSOD faces two main challenges: 1) reduce the interference of noisy background in complex scenes; 2) mine integral co-salient objects with large appearance variations. Some works introduce extra SOD dataset to provide saliency guidance~\cite{zhang2020coadnet, zhang2021summarize} or predict saliency maps~\cite{jin2020icnet} in order to mask out the co-salient objects. However, these approaches highly depend on the extra dataset, leading to supererogatory human effort to provide annotations.

Recent approaches~\cite{jin2020icnet, zhang2021summarize,fan2021group, zhang2021deepacg} try to use attention mechanism~\cite{vaswani2017attention} to strengthen co-salient features or build feature consistency to formulate the shared attributes of co-salient objects for integral predictions. However, there are two main drawbacks when directly applying attention mechanism for this task. On the one hand, the response maps reflecting the shared attributes, obtained in the attention mechanism, can only cover limited pixels belonging to co-salient objects, as shown in~\cref{FIG:cormap}.(b). In this case, it is difficult for the model to learn comprehensive shared attributes of co-salient objects. 
On the other hand, for complex scenes, the attention mechanism tends to focus on the wrong object regions, as shown in the second picture of\cref{FIG:cormap}.(b). Some methods such as GCoNet~\cite{fan2021group} propose a kind of group collaborative learning by collecting artificial negative group pairs. However, their pairs are grouped based on the auxiliary classification information, which requires major effort to group dissimilar negative category pairs as there is no clear definition of natural discrete object categories in real world~\cite{thoma2020soft}. 

To solve aforementioned issues, we design a novel \textbf{D}emocratic \textbf{C}o-salient-\textbf{F}eature-\textbf{M}ining framework (\textbf{DCFM}). Our DCFM can directly mine more comprehensive features and suppress the noisy background effectively without using extra SOD dataset or classification information. Specifically, in order to mine sufficient co-salient information, we first design a democratic prototype generation module (DPG), where democratic response maps are generated to capture more shared attributes. As shown in~\cref{FIG:cormap}.(c), our response maps cover more regions of co-salient objects. Then, a prototype with comprehensive co-salient information can be generated according to the democratic response maps, which can further guide the model to predict the co-salient objects.

Next, in order to suppress noisy background information in our prototype and avoid introducing extra classification information, we propose a simple self-contrastive learning module (SCL) to form positive and negative pairs to filter noise. We argue that the prototype generated from original images should be consistent with that generated when the image background regions are erased, and should be different from that generated when the co-salient objects are erased. Thus, a self-contrastive loss among these prototypes is designed to suppress the influence of noisy background and help the model learn more discriminative features of co-salient objects. 

Finally, to further strengthen the detected co-salient features from the above modules, we design a democratic feature enhancement module (DFE) based on the attention mechanism~\cite{vaswani2017attention}. As mentioned before, the attention mechanism tends to focus on a limited number of correlated features, which fails to provide comprehensive information. Therefore, we readjust the attention values to generate a  democratic attention map aggregating more correlated pixels for feature enhancement. 

Generally, our main contributions can be summarized as:
\begin{itemize}
	\item A democratic prototype generation module (DPG) is designed to build response maps covering sufficient co-salient regions, so as to generate a prototype containing comprehensive shared attributes as guidance for co-saliency prediction.
	\item A self-contrastive learning module (SCL) is proposed to help our model reduce the influence of noisy background without relying on additional classification information, where both positive and negative samples are generated from the image itself.
	\item A democratic feature enhancement module (DFE) is designed to further strengthen the co-salient features by adjusting attention values to involve more related pixels.
	\item Extensive experiments show that our method performs better than state-of-the-art methods, especially on challenging real-world cases, such as the CoCA dataset, we obtain a gain of 2.0\% for MAE, 5.4\% for maximum F-measure, 2.3\% for maximum E-measure, and 3.7\% for S-measure under the same settings.
\end{itemize}

\section{Related Work}
\subsection{Co-Salient Object Detection}
CoSOD is gaining increasing popularity. Some works establish graphs to model the relationship among pixels from a group of images~\cite{zhang2020adaptive, jiang2019unified, jiang2020co, hu2021multi, wei2019deep, jiang2019multiple}, then the co-salient objects can be mined with consistent features. Some works adopt extra salient object detection to mine salient objects first and then conduct CoSOD~\cite{zhang2020coadnet, jin2020icnet, zhang2021summarize}. Besides, SAEF~\cite{tsai2019deep} proposes to use saliency proposals generated by unsupervised deep learning based models first and then conduct CoSOD according to those proposals. Other works~\cite{fan2021group, zhang2020gradient, zhang2021deepacg} try to formulate shared attributes among input images to reflect the co-salient pixels and use classification information as a supplement of semantic information. In CoEGNet \cite{deng2021re}, edge detection is used for better structure prediction. More information on CoSOD can be found in surveys\cite{fan2020taking, zhang2018review, cong2018review}. Although these methods have obtained outstanding performances, they rely on extra information to learn discriminative co-salient features. Therefore, we consider thoroughly exploring the intrinsic characteristics of co-salient objects and background to realize CoSOD without using the SOD dataset or extra classification information.

\subsection{Contrastive Learning}
Contrastive learning is prevalent in self-supervised classification. SimCLR~\cite{chen2020simple} studies the importance of grouping negative pairs for contrastive learning. MoCo~\cite{he2020momentum} adopts a memory bank to reduce the influence of insufficient negative pairs. Additionally, contrastive learning has also be applied in many other tasks, such as video grounding~\cite{nan2021interventional}, long-tailed recognition~\cite{cui2021parametric}, action recognition~\cite{singh2021semi} and visual localization~\cite{thoma2020soft}. However, as argued in ~\cite{thoma2020soft}, it is difficult to define artificial negative pairs for contrastive learning, and they use soft assignments of images instead. In this paper, we also study how to group positive/negative pairs without artificial defined categories in CoSOD.

\subsection{Attention Mechanism}
Attention mechanism has been applied in different tasks. It has been applied in machine translation to draw global relationships between input and output~\cite{vaswani2017attention}. It is also utilized in non-local networks to deal with detection and segmentation~\cite{wang2018non, fu2019dual, wang2020few}. Besides, $A^2GNN$~\cite{zhang2021affinity} utilize it for affinity maps in weakly semantic segmentation. Additionally, it is used in video object segmentation to segment and track the target object~\cite{oh2019video,yu2021fast}. Moreover, it is deployed in referring expression grounding~\cite{sun2021discriminative} and language-person search~\cite{9406055}. Recently, attention mechanism has been applied in SOD or CoSOD.  
MSANet~\cite{zhou2020multi} applies attention mechanism to mine salient features and suppress background information. Besides, it is also used in CA-FCN~\cite{gao2020co} and RCAU~\cite{li2019detecting} as co-attention to link co-salient objects for CoSOD. However, we find that the attention mechanism tends to focus on a limited number of pixels. Therefore, in this paper, we try to introduce democracy into the attention mechanism to involve more related pixels.

\section{Methodology}
\begin{figure*}
	\centering
	\includegraphics[scale=0.295]{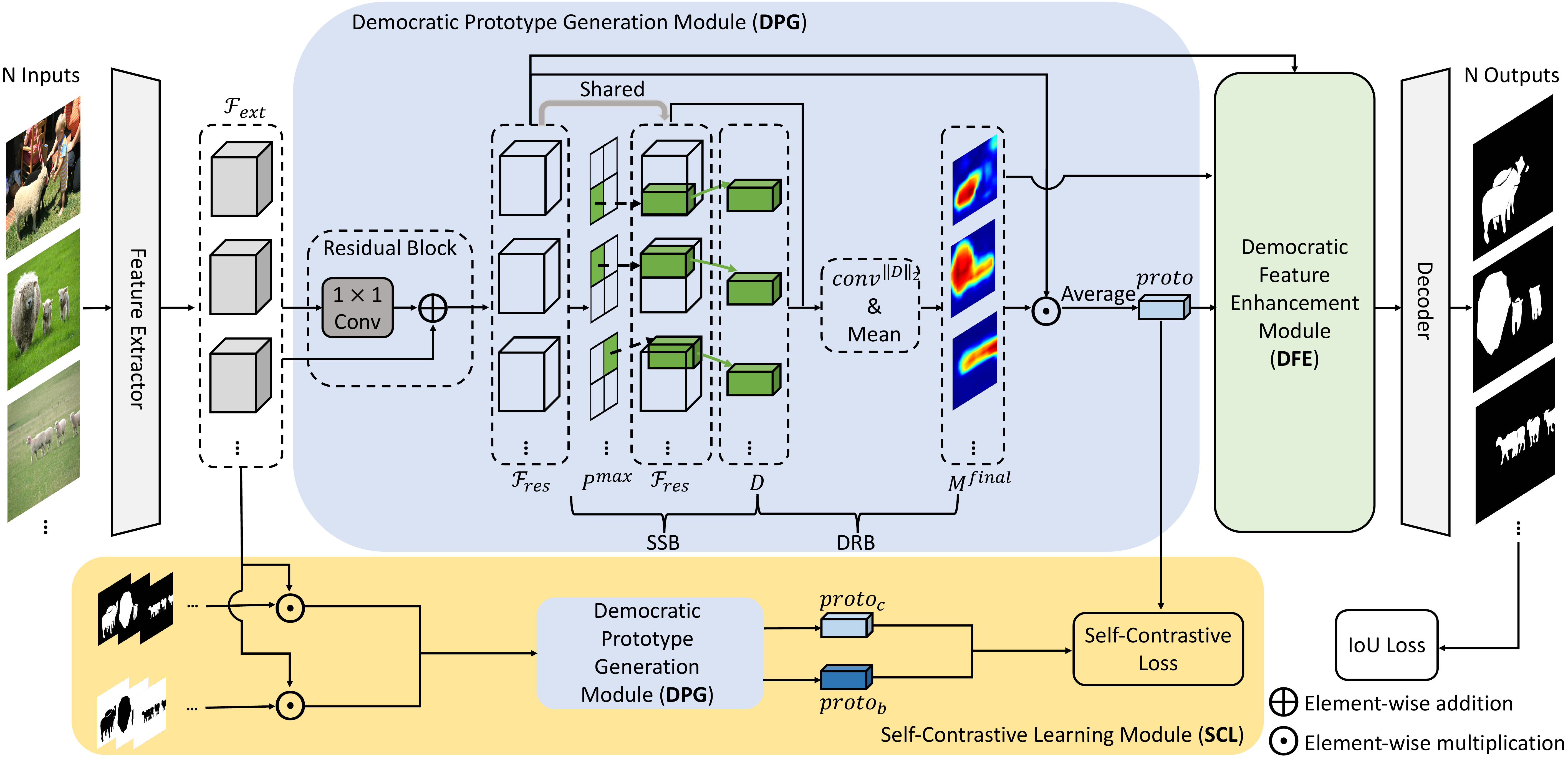}
	\caption{The framework of our network and the learning procedure. Specifically, the network contains five main parts, including a feature extractor, a democratic prototype generation module (DPG), a self-contrastive learning module (SCL), a democratic feature enhancement module (DFE), and a decoder. Note that the SCL is only used during training.}
	\label{FIG:overview}
\end{figure*}

\subsection{Overview}
The CoSOD dataset includes groups of images with labels. Each group is represented as $G=\{I, Y\}$, where $I=\{x_n\}_{n=1}^N$, $Y=\{y_n\}_{n=1}^N$, $x_n$ is the input image, $y_n$ is the corresponding label, $N$ is the total number of images in group $G$, and all images contain related objects. The labels are unavailable during inference. The model needs to detect the co-existed salient objects in each image of the same group. In this work, we aim to design a model that can detect the co-salient objects by thoroughly exploring the shared attributes to mine comprehensive co-salient features, and suppress noisy background through self-contrastive learning without using classification information or extra SOD dataset. 

The framework of our method and the learning procedure are demonstrated in~\cref{FIG:overview}. There are five main modules in our network, including a feature extractor, a democratic prototype generation module (DPG), a self-contrastive learning module (SCL), a democratic feature enhancement module (DFE), and a decoder. Note that the SCL is only applied for training and will be removed during inference. The overall process can be summarized as:

\begin{enumerate}[1)]
    \item Firstly, the feature extractor encodes a group of relative images ($N$ images) as initial features, which are then proceeded by the DPG to generate a comprehensive co-salient prototype. 
    \item Meanwhile, to avoid mining noisy information from the background in the prototype, our SCL is deployed for auxiliary training.
    \item Then, the prototype is fused into the visual features, and the fused features are transmitted into the DFE to strengthen the features further.
    \item Finally, the strengthened features are input into the decoder to predict the corresponding co-saliency maps.
\end{enumerate}

In the following sections, the details about the democratic prototype generation module, the self-contrastive learning module, and the democratic feature enhancement module will be discussed, respectively.

\subsection{Democratic Prototype Generation Module}
Our democratic prototype generation module (DPG) mainly contains three parts in series, which are the residual block, the seed selection block (SSB), and the democratic response block (DRB). 

After passing the feature extractor, we obtain the initial features $\mathcal{F}_{ext} \in \mathbb{R}^{N\times C \times H \times W}$ ($C$, $H$, $W$ are the channel number, height, and width), which are processed by the residual block first to generate strengthened residual features $\mathcal{F}_{res}$:
\begin{equation}
\centering
\mathcal{F}_{res} = \mathcal{F}_{ext} + conv^{1\times1}(\mathcal{F}_{ext}), 
\label{eq:residual}
\end{equation}
where $conv^{1\times1}$ represents for the $1\times1$ convolution layer and $\mathcal{F}_{res} \in \mathbb{R}^{N\times C \times H \times W}$. 

Then, the generated features $\mathcal{F}_{res}$ are passed into the SSB to select the most discriminative seeds for the co-salient objects in each input image. Next, the selected seeds are correlated with the residual feature maps to produce the response maps by the DRB. Finally, the response maps are multiplied with the residual features and averaged to generate the prototype, containing comprehensive co-salient feature information and guiding following prediction.

\begin{figure*}
	\centering
	\includegraphics[scale=0.218]{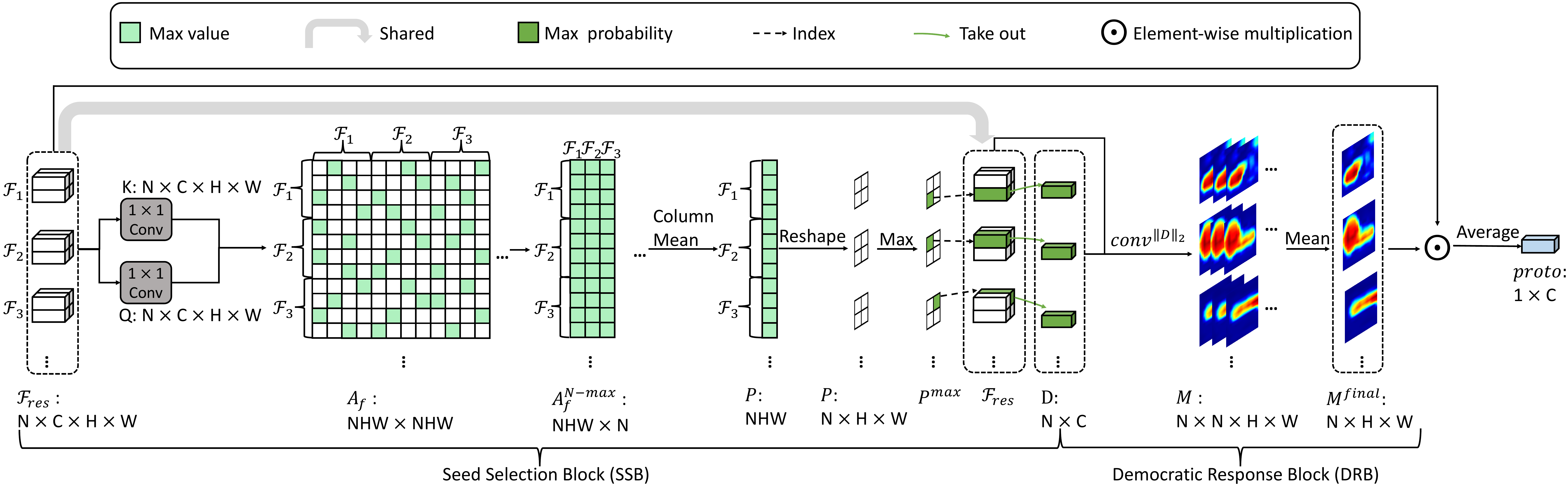}
	\caption{The framework of the seed selection block (SSB) and democratic response block (DRB). The inputs are the residual features. Then, the co-salient seeds are selected first from the residual features by SSB. After that, the response maps are produced using the selected seeds and the residual features through DRB. The final response maps and the input residual features are fused to generate the prototype.
	}
	\label{FIG:cssm}
\end{figure*}

\textbf{Seed Selection Block (SSB).} The SSB is demonstrated in~\cref{FIG:cssm}. This block is deployed to detect each image's most representative pixel as seed for response map generation. First, the residual features $\mathcal{F}_{res}$ are input to our SSB. Then, the attention mechanism is employed, in which two $1\times1$ convolution layers are deployed to obtain two feature maps, namely $K \in \mathbb{R}^{N\times C \times H \times W}$ and $Q \in \mathbb{R}^{N\times C \times H \times W}$. After reshaping both $K$ and $Q$ to shape $ \mathbb{R}^{NHW\times C}$, the feature similarity map ($S$) of each pixel is computed as
\begin{equation}
\centering
S = KQ^\top,
\label{eq:seedsatt}
\end{equation}
where $S\in \mathbb{R}^{NHW\times NHW}$, $\top$ means transpose,  and each row of $S$ represents similarities between one pixel and all pixels of the $N$ inputs.

Then, we first reshape $S$ into $S\in \mathbb{R}^{NHW\times N\times HW}$ and choose its maximum similarity value in each image, to get $N$ maximum similarity values for each pixel. This process is calculated by
\begin{equation}
\centering
S^{N\text{-}max} = \max_{i=1\cdots HW} S[:, :, i],
\label{eq:nmax}
\end{equation}
where $S^{N\text{-}max} \in \mathbb{R}^{NHW\times N}$. Afterwards, the average of the $N$ maximum similarity values is treated as the co-salient probability of each pixel,
\begin{equation}
\centering
P = \frac{1}{N} \sum_{n=1}^N S^{N\text{-}max}[:, n],
\label{eq:pmap}
\end{equation}
where $P \in \mathbb{R}^{NHW}$.

Then, the probability map is reshaped back to $P \in \mathbb{R}^{N \times H\times W}$. We can locate the pixel with the highest probability of being the co-salient object in each image by 
\begin{equation}
\centering
P^{max} = \max_{\begin{subarray}{l}
h=1,\cdots H  \\
w = 1,\cdots W
\end{subarray}} P[:, h, w],
\label{eq:seedsindex}
\end{equation}
\begin{equation}
\centering
\mathit{index} = ind(P^{max}),
\label{eq:seedsindex}
\end{equation}
where $ind(\cdot)$ means taking out the index of $P^{max}$. 

Finally, we take out the feature vectors from the $\mathcal{F}_{res}$ according to the index in~\cref{eq:seedsindex} as the final seeds by
\begin{equation}
\centering
D = \mathcal{F}_{res}(\mathit{index}).
\label{eq:seeds}
\end{equation}
Note that each image will provide one seed vector, and there are totally $N$ seeds. These seeds can represent the essential characteristics of the co-salient objects in each input image and be used for localization. 

\textbf{Democratic Response Block (DRB).} The DRB is demonstrated in~\cref{FIG:cssm}. If we directly use the seeds $D$ as the prototype, it fails to aggregate comprehensive characteristics of the co-salient objects. This is because it is difficult for limited seeds to express the integral co-existed objects, especially when there are large appearance variations among the group. Thus, we try to involve more pixels of the co-salient objects to generate a comprehensive prototype by considering the correlation between each pixel and the seeds $D$ from SSB.

Specifically, we first use L2 normalization in channel dimension to obtain the normalized residual features $\left \| \mathcal{F}_{res} \right \|_2$ and the normalized seeds $\left \| D \right \|_2$. Then, $\left \| D \right \|_2$ are treated as the kernel to conduct convolution on the $\left \| \mathcal{F}_{res} \right \|_2$: 
\begin{equation}
\centering
M= conv^{\left \| D \right \|_2}(\left \| \mathcal{F}_{res} \right \|_2), 
\label{eq:correlation}
\end{equation}
where $M$ means response maps, $conv^{\left \| D \right \|_2}$ is the convolution with $\left \| D \right \|_2$ as kernel. Since $D$ has $N$ seed vectors, the size of response maps become $\mathbb{R}^{N\times N \times H\times W}$ after~\cref{eq:correlation}, with channel dimension being the number of response maps for each input. 

The final democratic response map of each image is computed as the mean value of the $N$ response maps:
\begin{equation}
\centering
M^{\mathit{final}} = \frac{1}{N} \sum_{n=1}^{N} M[:, n, :, :],
\label{eq:correlationfinal}
\end{equation}
where $M^{final} \in \mathbb{R}^{N \times H\times W}$. In this way, more pixels have chance to contribute to the response maps.

Finally, the prototype ($proto\in \mathbb{R}^{1\times C}$) is generated by
\begin{equation}
\centering
\mathit{proto}=avg(M^{\mathit{final}}\odot \mathcal{F}_{res}),
\label{eq:proto}
\end{equation}
where the $M^{\mathit{final}}$ is broadcast to the same size as $\mathcal{F}_{res}$, $\odot$ denotes element-wise multiplication, and $avg(\cdot)$ means averaging feature vector of all the pixels from all inputs.

\subsection{Self-Contrastive Learning Module}
To further help the DPG to suppress the noise of background, and learn discriminative features without depending on classification information, a self-contrastive learning module (SCL) is designed as shown in~\cref{FIG:overview}. Our motivation is that the prototype generated by the original inputs ($\mathit{proto}$) should be consistent with co-salient prototype generated by inputs where background is erased ($\mathit{proto}_c$), but different from the background prototype generated by inputs where the co-salient objects are erased ($\mathit{proto}_b$). Note that the inputs here are the initial extracted features $\mathcal{F}_{ext}$ from the feature extractor. The co-salient prototype and background prototype can be generated as
\begin{equation}
\centering
 \mathit{proto}_c = \phi_{\text{DPG}}(\mathcal{F}_{ext}  \odot   Y^{\downarrow}), 
\label{eq:protoc}
\end{equation}
\begin{equation}
\centering
 \mathit{proto}_b = \phi_{\text{DPG}}(\mathcal{F}_{ext}  \odot (1- Y^{\downarrow})), 
\label{eq:protob}
\end{equation}
where $\phi_{\text{DPG}}$ is short for the process of DPG, `$\downarrow$' means downscaling the groundtruth $Y$ to the same size as $\mathcal{F}_{ext}$ then broadcasting to the same channel number.

Then, $\mathit{proto}$ and $\mathit{proto}_c$ are treated as a positive pair, while $\mathit{proto}$ and $\mathit{proto}_b$ are treated as a negative pair. A self-contrastive loss is designed to pull together the positive pair and push away the negative pair. First, we define the cosine-style similarity between the prototypes by
\begin{equation}
\centering
cos(p_1, p_2) =(1+{p_1 \cdot  p_2 \over  \left | p_1 \right | \left | p_2  \right | })\times 0.5.
\label{eq:cosine}
\end{equation}
After that, the self-contrastive loss is defined as
\begin{equation}
\centering
 cos_c = cos(\mathit{proto}, \mathit{proto}_c), 
\label{eq:cosc}
\end{equation}
\begin{equation}
\centering
 cos_b = cos(\mathit{proto}, \mathit{proto}_b), 
\label{eq:cosb}
\end{equation}
\begin{equation}
\centering
\mathcal{L}_{sc} = - log(cos_c + \epsilon)-log(1 - cos_b+\epsilon),
\label{eq:scl}
\end{equation}
where $\epsilon$ is a small constant value ensuring non-zero values for $log(\cdot)$ and set as $1\times 10^{-5}$. Our SCL is only applied during training as an auxiliary loss to help the DPG learn more discriminative co-salient features. This part is not used during inference.

\subsection{Democratic Feature Enhancement Module}
We design a democratic feature enhancement module (DFE) to further strengthen the fused co-salient features from DPG for final prediction. Our DFE is based on the attention mechanism~\cite{vaswani2017attention}. We observe that conventional attention~\cite{vaswani2017attention} tends to focus on a limited number of related pixels. Thus, we argue that democracy also matters in this case, and more pixels should be involved in enhancing the fused features. Thus, we try to amplify small positive attention values to involve more pixels for feature enhancement. Negative attention values are not considered here as they usually represent irrelevance.
\begin{figure}
\centering
	\includegraphics[scale=0.26]{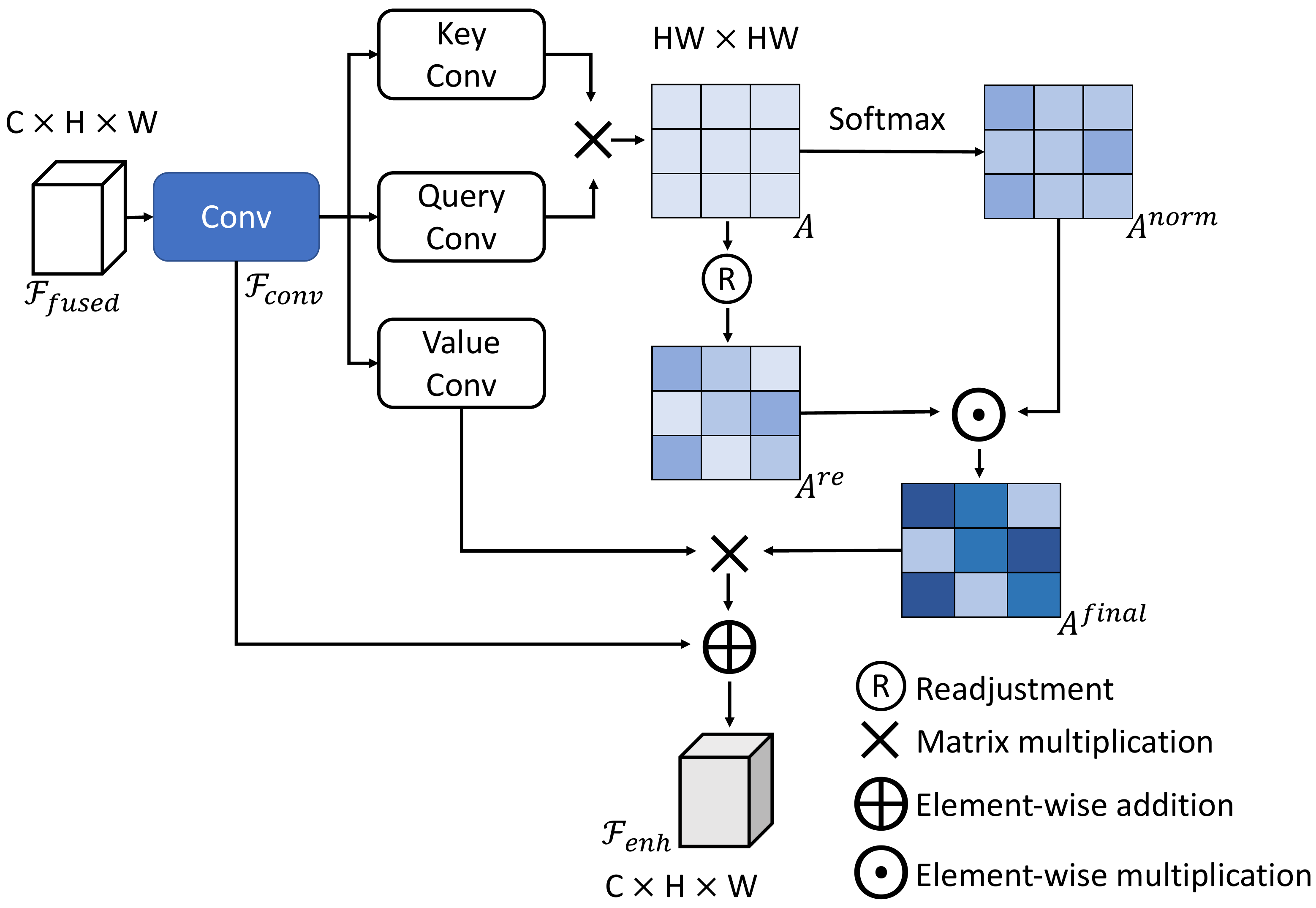}
	\caption{Flow chat of democratic feature enhancement module.}
	\label{FIG:attention}
\end{figure}
First, we generate the fused features using the guidance of both response maps and the prototype derived from~\cref{eq:correlationfinal} and~\cref{eq:proto} in the DPG as
\begin{equation}
\centering
\mathcal{F}_{\mathit{fused}} = \mathcal{F}_{res} \odot M^{\mathit{final}} + \mathcal{F}_{res} \odot  proto,
\label{eq:outputfeature}
\end{equation}
where both the $M^{\mathit{final}}$ and $\mathit{proto}$ are broadcast into the same size as $\mathcal{F}_{res}$. Therefore, the fused features by~\cref{eq:outputfeature} contain both specific attributes and shared attributes.

The fused features of each input image are enhanced with our DFE individually and independently. As shown in~\cref{FIG:attention}, the corresponding $\mathcal{F}_{\mathit{fused}}$ is input to a $1\times 1$ convolution layer followed by a ReLU activation to obtain $\mathcal{F}_{conv}\in \mathbb{R}^{C\times H \times W}$ first. After that, key, query and value convolutions are applied and then reshaped to generate $\mathcal{F}_k \in \mathbb{R}^{HW\times C}$, $\mathcal{F}_q \in \mathbb{R}^{HW\times C}$ and $\mathcal{F}_v \in \mathbb{R}^{ HW\times C}$. Then, the initial attention map ($A$) can be computed by 
\begin{equation}
\centering
A = \mathcal{F}_k \mathcal{F}_q^\top,
\label{eq:dematt}
\end{equation}
where $A \in \mathbb{R}^{HW\times HW}$ and $\top$ means transpose. 

Next, a softmax is applied to $A$ to obtain the normalized attention map ($A^{norm}$). Moreover, the initial attention map $A$ is sorted in a descending order to generate the sorting index matrix ($Z$). As we adopt the descending order, the small attention values are assigned with large sorting index. Then, we apply the following formula to amplify the small positive attention values,
\begin{equation}
\centering
A^{re}_{i,j}=\left\{\begin{matrix} 
  (Z_{i,j}+1)^{\alpha}, \text{if } A_{i, j}>0 \\ 
  1, \text{else}
\end{matrix}\right.,
\label{eq:reatt}
\end{equation}
where $A^{re}$ denotes the weights for readjusting the attention, $\alpha$ is a coefficient for determining the degree of amplification, $i$ and $j$ are the spatial index. Then, the final attention map is computed by
\begin{equation}
\centering
A^{\mathit{final}} = A^{norm} \odot A^{re},
\label{eq:finalatt}
\end{equation}
and the final enhanced features can be computed by
\begin{equation}
\centering
\mathcal{F}_{enh} = \mathcal{F}_{conv} + A^{\mathit{final}} \mathcal{F}_{v},
\label{eq:augfeature}
\end{equation}
where the result of $A^{\mathit{final}} \mathcal{F}_{v}$ is first reshaped back into the same size as $\mathcal{F}_{conv}$. Finally, the augmented features $\mathcal{F}_{enh}$ are transmitted into the decoder to predict the corresponding co-saliency maps. 

\subsection{Objective Function}
The objective function for training is a combination of IoU loss~\cite{qin2019basnet, zhang2020coadnet} and our self-contrastive loss in~\cref{eq:scl}. The IoU loss can be illustrated as
\begin{equation}
\centering
\mathcal{L}_{iou} = 1 - \frac{1}{N}\sum\frac{\hat{Y} \cap Y}{\hat{Y} \cup Y},
\label{eq:iou}
\end{equation}
where $\hat{Y}$ denotes for predictions and $Y$ denotes for the groundtruth. Then, the final objective function is
\begin{equation}
\centering
\mathcal{L}_{tot} = \mathcal{L}_{iou} + \lambda \mathcal{L}_{sc},
\label{eq:loss}
\end{equation}
where $\lambda$ is to balance IoU loss and self-contrastive loss.

\begin{table*}
		\centering
		\caption{Comparisons with other state-of-the-art approaches on 3 benchmarks. $\uparrow$ means that larger is better and $\downarrow$ denotes that smaller is better. `SOD' denotes training with extra SOD dataset. 
}
	\scalebox{0.75}{
	\begin{tabular}{c|c|cccc|cccc|cccc}
		\bottomrule
		& & \multicolumn{4}{| c |}{CoCA} &\multicolumn{4}{| c |}{CoSOD3k}&\multicolumn{3}{| c }{Cosal2015} \\
		Methods & SOD & $\mathit{MAE}\downarrow$ & $F_\beta^{max} \uparrow$ & $E_\xi^{max} \uparrow$ & $S_\alpha \uparrow$ & $\mathit{MAE}\downarrow$ & $F_\beta^{max} \uparrow$ & $E_\xi^{max} \uparrow$ & $S_\alpha \uparrow$ & $\mathit{MAE}\downarrow$ & $F_\beta^{max} \uparrow$ & $E_\xi^{max} \uparrow$ & $S_\alpha \uparrow$  \\
		\midrule
		ICNet~\cite{jin2020icnet} (NeurIPS20) & \checkmark & 0.148 & 0.506 & 0.698 & 0.651 & 0.097 & 0.744 & 0.832 & 0.780 & 0.058 & 0.855 & 0.900 & 0.856\\
		CADC~\cite{zhang2021summarize} (ICCV21) & \checkmark & 0.132 & 0.548 & 0.744 & 0.681 & 0.096 & 0.759 & 0.840 & 0.801 & 0.064 & 0.862 & 0.906 & 0.866\\
		CoADNet~\cite{zhang2020coadnet} (NeurIPS20) & \checkmark & - & - & - & - & 0.070 & 0.825 & - & 0.837 & 0.064 & 0.875 & - & 0.861 \\ 
		\midrule
		CSMG~\cite{zhang2019co} (CVPR19) &  & 0.124 & 0.503 & 0.734 & 0.632 & 0.157 & 0.645 & 0.723 & 0.711 & 0.130 & 0.777 & 0.818 & 0.774\\
		GCAGC~\cite{zhang2020adaptive} (CVPR20) &  & 0.111 & 0.523 & 0.754 & 0.669 & 0.100 & 0.740 & 0.816 & 0.785 & 0.085 & 0.813 & 0.866 & 0.817\\
		CoEGNet~\cite{deng2021re} (TPAMI21) &  & 0.106 & 0.493 & 0.717 & 0.612 & 0.092 & 0.736 & 0.825 & 0.762 & 0.077 & 0.832 & 0.882 & 0.836 \\
		GICD~\cite{zhang2020gradient} (ECCV20)&  & 0.126 & 0.513 & 0.715 & 0.658  & 0.079 & 0.770 & 0.848 & 0.797 & 0.071 & 0.844 & 0.887 & 0.844 \\
		DeepACG~\cite{zhang2021deepacg} (CVPR21) &  & 0.102 & 0.552 & 0.771 & 0.688 & 0.089 & 0.756 & 0.838 & 0.792 & \textbf{0.064} & 0.842 & 0.892 & \textbf{0.854} \\
		GCoNet~\cite{fan2021group} (CVPR21) &  & 0.105 & 0.544 & 0.760 & 0.673 & 0.071 & 0.777 & 0.860 & 0.802 & 0.068 & 0.847 & 0.887 & 0.845 \\
		DCFM (Ours) &  & \textbf{0.085} & \textbf{0.598} & \textbf{0.783} & \textbf{0.710} & \textbf{0.067} & \textbf{0.805} & \textbf{0.874} & \textbf{0.810} & 0.067  & \textbf{0.856} & \textbf{0.892} & 0.838\\
		\bottomrule
	\end{tabular}}
\label{tbl:overall}
\end{table*}

\begin{figure*}
	\centering
	\includegraphics[scale=0.225]{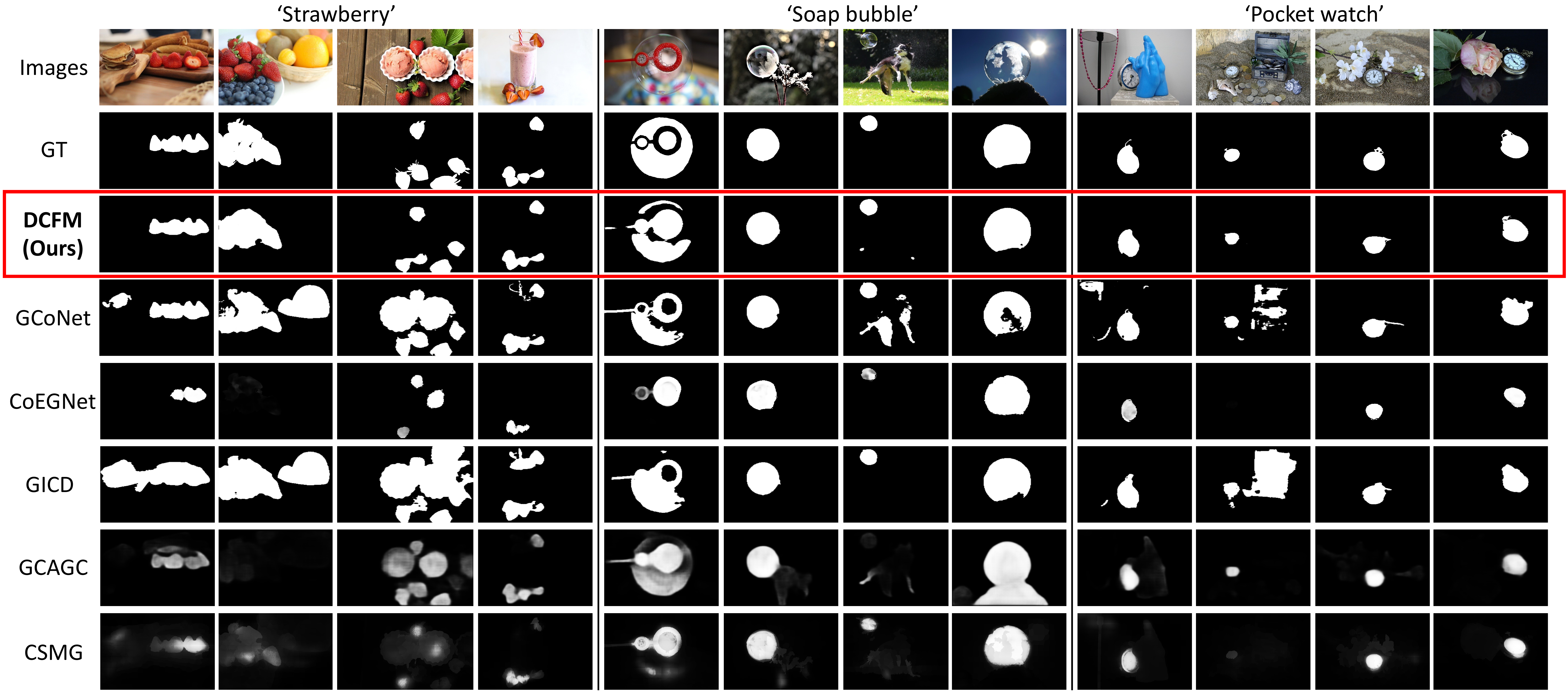}
	\caption{The qualitative comparisons with other state-of-the-art methods. It is evident that our method can predict smoother co-saliency maps with less noise of background, compared with other state-of-the-art methods. More can be found in our supplementary material.}
	\label{FIG:qualitative}
\end{figure*}
\section{Experiment}
\begin{table*}
	\centering
	\caption{Ablation study for our proposed modules. `Base.' denotes baseline. Our overall method obtains the best results.}
	\begin{tabular}{c|c|c|c|c|cccc}
		\bottomrule
		& Base. & DPG & SCL & DFE & $\mathit{MAE}\downarrow$ & $F_\beta^{max} \uparrow$ & $E_\xi^{max} \uparrow$ & $S_\alpha \uparrow$ \\
		\midrule
		1 & \checkmark &   &   &   & 0.129 & 0.521 & 0.735 & 0.655 \\
		2 & \checkmark & \checkmark &   &  & 0.097 & 0.575 & 0.763 & 0.696 \\
		3 & \checkmark & \checkmark & \checkmark & & 0.087 & 0.592 & 0.775 & 0.701 \\
		4 & \checkmark & \checkmark & \checkmark & \checkmark & \textbf{0.085} & \textbf{0.598} & \textbf{0.783} & \textbf{0.710} \\
		\bottomrule
	\end{tabular}
\label{tbl:ablation}
\end{table*}
\begin{table}
	\centering
	\caption{Ablation study for different parts in DPG. `RB' means the residual block. The overall process obtains the best performance.}
	\scalebox{0.81}{
	\begin{tabular}{c|c|c|c|cccc}
		\bottomrule
		& RB & SSB & DRB & $\mathit{MAE}\downarrow$ & $F_\beta^{max} \uparrow$ & $E_\xi^{max} \uparrow$ & $S_\alpha \uparrow$ \\
		\midrule
		1 &  &  &  & 0.129 & 0.521 & 0.735 & 0.655 \\
		2 & \checkmark & & & 0.124 & 0.527 & 0.745 & 0.659 \\
		3 & \checkmark & \checkmark &  & 0.126 & 0.527 & 0.739 & 0.657 \\
		4 & \checkmark & \checkmark & \checkmark & \textbf{0.097} & \textbf{0.575} & \textbf{0.763} & \textbf{0.696} \\
		\bottomrule
	\end{tabular}
	}
\label{tbl:dpgmablation}
\end{table}

\subsection{Implementation Details}
We use Feature Pyramid Network (FPN)~\cite{lin2017feature} with VGG-16~\cite{simonyan2014very} as our backbone. The hyper-parameter $\alpha$ in~\cref{eq:reatt} is 3 and $\lambda$ in~\cref{eq:loss} is 0.1. Additionally, we use Adam~\cite{kingma2014adam} as our optimizer to train our model for 200 epochs. The learning rate is set as $1 \times 10^{-5}$ for feature extractor and $1 \times 10^{-4}$ for other parts. The weight decay is set as $1 \times 10^{-4}$. In each training episode, we randomly choose one group (16 samples) of relative images. For inference, all samples in each group are input at one time. The inputs are resized into $224\times 224$ for both training and inference. The computational complexity of~\cref{eq:seedsatt} is $O((NHW)^{2})$ and~\cref{eq:dematt} is $O((HW)^{2})$. The increment of FLOPs is small since the input size is small. The total training time is around 3 hours and the inference time is around 84.4 fps. All experiments are run on one NVIDIA GeForce RTX 2080 Ti.

\subsection{Dataset and Evaluation Metrics}
\textbf{Dataset.} We use COCO-SEG~\cite{wang2019robust}, a subset of COCO dataset~\cite{lin2014microsoft}, which contains 9,213 images from 65 groups for training. We evaluate our method on three popular CoSOD benchmarks: CoCA~\cite{zhang2020gradient}, Cosal2015~\cite{zhang2016detection} and CoSOD3k~\cite{fan2020taking}. CoCA and CoSOD3k are proposed for challenging real-world co-saliency evaluation, containing multiple co-salient objects in some images, large appearance and scale variations, and complex background clutters. Cosal2015 is a widely used large dataset for the evaluation.

\textbf{Evaluation Metrics.} The evaluation metrics include mean absolute error ($\mathit{MAE}$)~\cite{cheng2013efficient}, maximum F-measure ($F_{\beta}^{max}$)~\cite{achanta2009frequency}, maximum E-measure ({$E_{\phi}^{max}$})~\cite{fan2018enhanced} and S-measure ($S_\alpha$)~\cite{fan2017structure}. Specifically, the value of $\mathit{MAE}$ is the smaller, the better. While others are the larger, the better.

\subsection{Comparison with State-of-The-Art}
\textbf{Compared Methods.} We mainly compare with previous state-of-the-art methods trained on common single CoSOD training dataset for fair comparison, including CSMG~\cite{zhang2019co}, GCAGC~\cite{zhang2020adaptive}, CoEGNet~\cite{deng2021re}, GICD~\cite{zhang2020gradient}, GCoNet~\cite{fan2021group}, and DeepACG~\cite{zhang2021deepacg}. We also list several methods trained on both CoSOD dataset and SOD dataset, such as CADC~\cite{zhang2021summarize}, ICNet~\cite{jin2020icnet} and CoADNet~\cite{zhang2020coadnet}.

\textbf{Quantitative Comparison.} In~\cref{tbl:overall}, we list the performance comparisons between ours and previous state-of-the-art methods. It can be seen that our method reaches a new state-of-the-art performance compared with other approaches under the same settings. Specifically, for the two challenging real-world datasets CoCA and CoSOD3k, \eg, for CoCA, we obtain a gain of 2.0\% for MAE, 5.4\% for maximum F-measure, 2.3\% for maximum E-measure, and 3.7\% for S-measure compared with GCoNet~\cite{fan2021group}. Moreover, our method can even outperform those trained with extra SOD dataset on these two datasets, such as ICNet~\cite{jin2020icnet} and CADC~\cite{zhang2021summarize}. 
For Cosal2015, our method obtains comparable results with DeepACG~\cite{zhang2021deepacg} and GCoNet~\cite{fan2021group}. This phenomenon may be caused by the fact that both DeepACG~\cite{zhang2021deepacg} and GCoNet~\cite{fan2021group} use extra classification information to provide structure information, while our method does not rely on any extra information.

\textbf{Qualitative Comparison.} We also report some qualitative comparisons with state-of-the-art methods in~\cref{FIG:qualitative}. The groups are from CoCA dataset. It can be found that our model can predict more integral and less noisy co-saliency maps compared with others. Specifically, when there are multiple co-salient objects in one image, like the group `Strawberry', our model can detect all the target objects,  compared with CoEGNet~\cite{deng2021re} and CSMG~\cite{zhang2019co}. In the group `Soap bubble', ours are sensitive to appearance variations compared with others. When the background noise level is high, such as the group `Pocket watch', our predictions contain less noise, compared with GCoNet~\cite{fan2021group} and GICD~\cite{zhang2020gradient}.

\subsection{Ablation Study}

We conduct the ablation study of our method on the CoCA dataset by adding one module each time and treating the network with all our modules removed as the baseline. The results are shown in~\cref{tbl:ablation}. It can be found that each proposed module contributes a lot. With our DPG, the performance can increase 3.2\% for MAE, 5.4\% for maximum F-measure, 2.8\% for maximum E-measure, and 4.1\% for S-measure. Our SCL enables further improvement by 1.0\% for MAE, 1.7\% for maximum F-measure, 1.2\% for maximum E-measure, and 0.5\% for S-measure, respectively. Besides, our model with DFE reach 0.085 for MAE, 0.598 for maximum F-measure, 0.783 for maximum E-measure, and 0.710 for S-measure. The new state-of-the-art performance is obtained when all the modules are included.

\begin{table}
	\centering
	\caption{Ablation study for different parts in~\cref{eq:scl} of SCL. `$cos_c$' denotes the case with only positive pair for the loss and `$cos_b$' denotes that with only negative pair. DFE is not used. }
	\scalebox{0.95}{
	\begin{tabular}{c|c|c|cccc}
		\bottomrule
		& $cos_c$ & $cos_b$ & $\mathit{MAE}\downarrow$ & $F_\beta^{max} \uparrow$ & $E_\xi^{max} \uparrow$ & $S_\alpha \uparrow$ \\
		\midrule
		1 & & & 0.097 & 0.575 & 0.763 & 0.696 \\
		2 & \checkmark &   & 0.093 & 0.574 & 0.764 & 0.695 \\
		3 &  & \checkmark & 0.095 & 0.583 & 0.773 & 0.697 \\
		4 & \checkmark & \checkmark & \textbf{0.087} & \textbf{0.592} & \textbf{0.775} & \textbf{0.701} \\
		\bottomrule
	\end{tabular}
	}
\label{tbl:scablation}
\end{table}

\textbf{Impact of Democratic Prototype Generation Module.} The evaluation of each block of our DPG is listed  in~\cref{tbl:dpgmablation}. The experiment is conducted by adding one block at a time. Compared with the baseline (row 1), each part of DPG devotes to the final results. Specifically, if we only use RB and SSB, where we take the mean of seeds as the prototype, the results are even lower than the case without SSB, comparing row 2 and 3. On the other hand, with DRB, comparing row 3 and 4, the results will be increased by 2.9\% for MAE, 4.8\% for maximum F-measure, 2.4\% for maximum E-measure, and 3.9\% for S-measure. This phenomenon can verify that democracy does matter. More co-salient pixels should be enrolled for the comprehensive prototype.

\begin{table}
	\centering
	\caption{Ablation study for readjustment in DFE. `w/o DFE' denotes not using DFE, `w/o RA' denotes using DFE without readjustment and `w/ RA' denotes using DFE with readjustment.}
	\begin{tabular}{c|c|c|ccccc}
		\bottomrule
		&  & $\mathit{MAE}\downarrow$ & $F_\beta^{max} \uparrow$ & $E_\xi^{max} \uparrow$ & $S_\alpha \uparrow$ \\
		\midrule
		1 & w/o DFE & 0.087 & 0.592 & 0.775 & 0.701 \\
		2 & w/o RA & 0.100 & 0.567 & 0.769 & 0.691 \\
		3 & w/ RA & \textbf{0.085} & \textbf{0.598} & \textbf{0.783} & \textbf{0.710} \\
		\bottomrule
	\end{tabular}
\label{tbl:demablation}
\end{table}

\textbf{Impact of Self-Contrastive Learning Module.} We also evaluate two main parts in our self-contrastive loss as listed in~\cref{tbl:scablation}. We conduct this experiment by removing one part each time. It can be seen that with only positive pair $cos_c$, by comparing row 1 and 2, we can get comparable results. With only negative pair $cos_b$, the performance is clearly improved, by comparing row 1 and 3. This phenomenon proves that the negative pair is important for removing background noise. Nevertheless, by comparing row 3 and 4, the contrastive learning with both positive and negative pairs promotes balanced training for higher results. More analysis can be found in our supplementary material.

\textbf{Impact of Democratic Feature Enhancement Module.} We also experiment on the readjustment of attention values in~\cref{tbl:demablation}. When the readjustment is removed but using conventional attention in our DFE, the performance is even worse than the case without our DFE, as shown in row 1 and 2. Thus, democracy does matter in this module as well. Conventional attention mechanism focusing on limited pixels cannot provide sufficient information for the decoder while more related pixels should be involved.

\section{Conclusion}
In this paper, we have proposed a new method for CoSOD without using the SOD dataset and classification information. We design a democratic prototype generation module (DPG) to build democratic response maps first so as to generate a comprehensive prototype as guidance for further prediction. Moreover, to help suppress noisy background information in the prototype, we design a self-contrastive learning module (SCL), where both positive and negative pairs are generated from the image itself without relying on classification information. Besides, we also design a democratic feature enhancement module (DFE) to strengthen co-salient features from DPG for final prediction. Both our DPG and DFE show that democracy does matter. More related pixels should be involved for mining comprehensive features for CoSOD.

{\small
\bibliographystyle{ieee_fullname}
\bibliography{reference}
}
\clearpage

\begin{figure*}
\centering
	\includegraphics[scale=0.24]{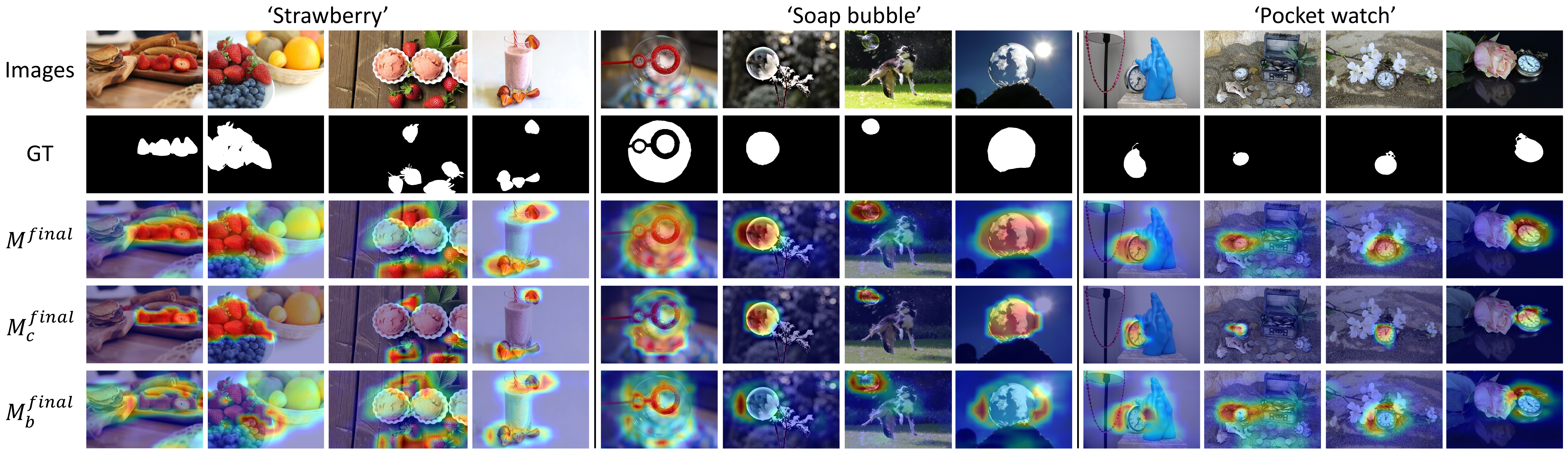}
	\captionof{figure}{Visualization of the response maps in different cases. The visualizations can verify our assumption of the self-contrastive learning module as $M^{\mathit{final}}$ is consistent with $M^{\mathit{final}}_c$ but different from $M^{\mathit{final}}_b$.}
	\label{FIG:scl}
\end{figure*}%

\appendix
\begin{figure*}
\centering
	\includegraphics[scale=0.23]{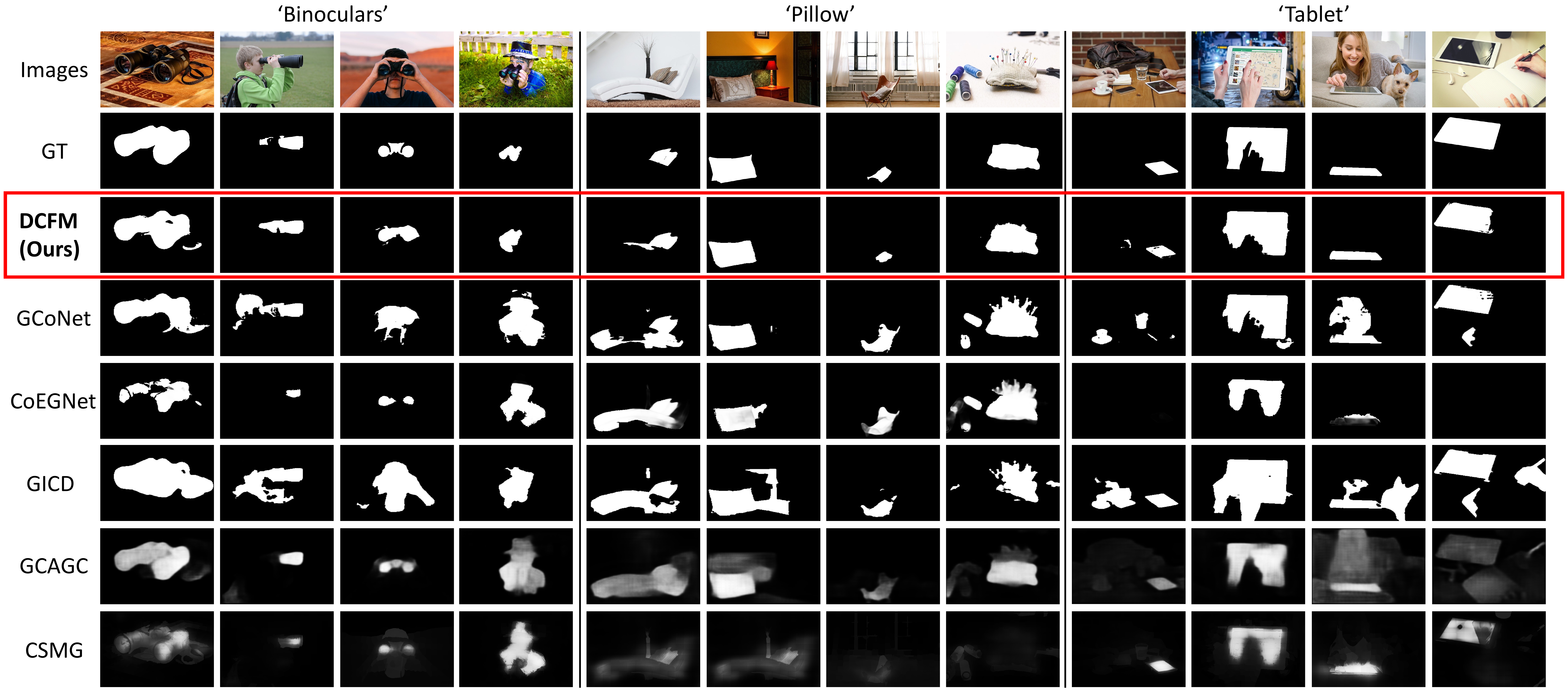}
	\caption{More visualizations of our predictions and comparisons with previous state-of-the-art approaches. It can be found that our model can better differentiate the co-salient objects and background in complex scenes.}
	\label{FIG:qual}
\end{figure*}
\section{Supplementary Material}
In this supplementary material, we will analyze the self-contrastive learning module (SCL) with some visualizations. Besides, we will provide more qualitative comparisons between our model and other state-of-the-art approaches. Additionally, we will compare the complexity of our method with others. We will also analyze the influence of $\alpha$ in Eq.(19) in our paper. Moreover, we will discuss about some failure cases.

\subsection{Self-Contrastive Learning Module Analysis}
We display some response maps in different cases on the CoCA dataset~\cite{zhang2020gradient} in~\cref{FIG:scl}. Note that this dataset is used for evaluation. $M^{\mathit{final}}$ denotes the normal response maps generated by original inputs, $M^{\mathit{final}}_c$ denotes the co-salient response maps generated by inputs where the background regions are erased, and $M^{\mathit{final}}_b$ denotes the background response maps generated by inputs where the co-salient objects are erased. Then, $\mathit{proto}$, $\mathit{proto}_c$ and $\mathit{proto}_b$ can be derived based on the corresponding response maps. As shown in~\cref{FIG:scl}, it can be found that the $M^{\mathit{final}}$ can focus on most regions of the target co-salient objects. Moreover, comparing $M^{\mathit{final}}_c$ and $M^{\mathit{final}}_b$, the $M^{\mathit{final}}_c$ can highlight all the related co-salient objects. In contrast, the $M^{\mathit{final}}_b$ are sensitive to the surroundings of the co-salient objects. In this case, our assumption of SCL, where $\mathit{proto}$ and $\mathit{proto}_c$ are pulled together while $\mathit{proto}$ and $\mathit{proto}_b$ are pushed away, can be verified. With our SCL, the model can learn to differentiate co-salient features and background features. Thus, the noise information can be suppressed.

\subsection{Qualitative Comparison}
We list more qualitative comparisons with previous sate-of-the-art methods in~\cref{FIG:qual}. We use the CoCA dataset~\cite{zhang2020gradient} for demonstration, as it is a challenging real-world dataset, containing more challenging cases. The compared methods include CSMG~\cite{zhang2019co}, GCAGC~\cite{zhang2020adaptive}, CoEGNet~\cite{deng2021re}, GICD~\cite{zhang2020gradient}, GCoNet~\cite{fan2021group}, and DeepACG~\cite{zhang2021deepacg}. It is evident that our predictions are closer to the ground truth. Specifically, when the background contains misleading objects, such as the humans in the group `Binoculars', our model can suppress the noisy information and focus on the targets, compared with GCoNet~\cite{fan2021group} and GICD~\cite{zhang2020gradient}. Additionally, when there are complex background clutters, like images in the groups `Pillow' and `Tablet', compared with all other methods, ours are robust to this challenging setting.

\subsection{Complexity Analysis with State-of-the-art Methods} The computational complexity of Eq.(2) and Eq.(18) in our paper  is $O((NHW)^{2})$ and $O((HW)^{2})$ respectively. The increment of FLOPs is small since the input size is small. We list the complexity comparisons in~\cref{tbl:complexity}, `$\dag$' means without DFE. Ours can achieve an impressive performance with fewer FLOPs and parameters compared with CADC~\cite{zhang2021summarize} and GICD~\cite{zhang2020gradient}. Besides, ours can obtain a better performance with limited increment of FLOPs and parameters compared with GCoNet~\cite{fan2021group}, especially for DCFM$\dag$. 
Overall, our method has an impressive performance with comparable runtime.
\begin{table}
	\centering
	\caption{Complexity comparisons. `param.' denotes the number of parameters. We set 5 inputs to compute FLOPs. }
	\scalebox{0.74}{
	\begin{tabular}{c|c|c|c|c}
		\bottomrule
		method & FLOPs (G) & param. (M) & runtime (fps) & $F_\beta^{max} \uparrow$ \\
		\midrule
		CADC~\cite{zhang2021summarize}$_\text{ICCV21}$& 457.9 & 392.8 & 18.0 & 0.548 \\
		GICD~\cite{zhang2020gradient}$_\text{ECCV20}$& 467.6 & 278.0 & 40.8 & 0.513 \\
		GCoNet~\cite{fan2021group}$_\text{CVPR21}$ & 311.5 & 142.0 & 116.2 & 0.544 \\
		DCFM\dag (ours) & 313.0 & 140.5 & 101.9 & 0.592 \\
		DCFM (ours) & 316.6 & 142.3 & 84.4 & \textbf{0.598} \\
		\bottomrule
	\end{tabular}
	}
\label{tbl:complexity}
\end{table}

\begin{table}
	\centering
	\caption{Influence of alpha in Eq.(19) in our paper.}
	\scalebox{0.85}{
	\begin{tabular}{c|c|c|c|c|c}
		\bottomrule
		 $\alpha$ & 0.1 & 1 & 2 & 3 & 4 \\
		\midrule
		$F_\beta^{max} \uparrow$ & 0.578 & 0.592 & 0.593 & \textbf{0.598} & 0.587 \\
		\bottomrule
	\end{tabular}}
\label{tbl:alpha}
\end{table}

\begin{figure}
\centering
	\includegraphics[scale=0.43]{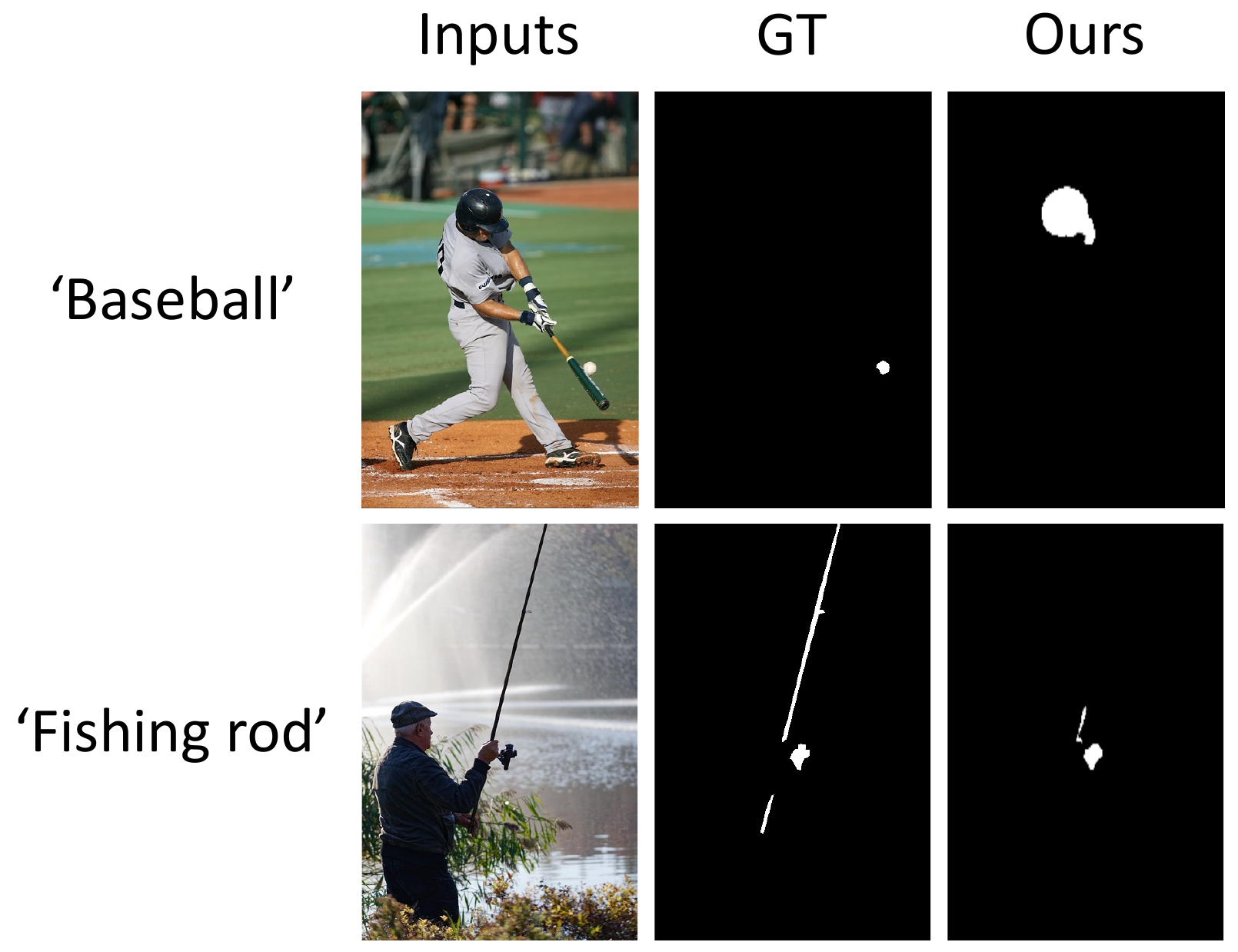}
	\caption{Visualizations of some failed cases.}
	\label{FIG:failure}
\end{figure}

\subsection{Influence of Alpha in Eq.(19) in Our Paper} We add the ablation study of alpha in~\cref{tbl:alpha}. The performance smoothly increases with larger alpha. However, performance decreases when alpha is too big ($\alpha$=4). When $\alpha \text{\textgreater}$4, the model even fails to be trained. This is because in this case, the weight of small positive attention values will be much bigger. Thus, the attention mechanism will be confused and tend to focus on those small values but neglect original high values.

\subsection{Limitation Discussion}
We also report some failure cases in~\cref{FIG:failure}. 
As shown in the figure, it is difficult for our model to predict small objects precisely. This may be caused by the fact that the inputs are resized into the size of $224\times 224$. Then, with the feature extractor, the size of the output features is $14\times 14$. In this case, it may cause information lost for small objects. Thus, it is difficult for our model to capture the corresponding features. Therefore, how to enhance model robustness for small objects is a direction for our future work.

\end{document}